\pdfoutput=1
\documentclass[11pt]{article}
\usepackage[final]{acl}
\newif\ifcamera
\cameratrue
\usepackage[T1]{fontenc}
\usepackage[utf8]{inputenc}
\usepackage{newtxtext}
\usepackage[varqu,varl]{inconsolata}

\usepackage{microtype}
\usepackage{graphicx}
\usepackage{booktabs}
\usepackage{array}
\usepackage{amssymb}

\newcommand{\arimg}[1]{\raisebox{-0.62ex}{\includegraphics[height=2.7ex]{ar/#1}}}

\newcommand{\ci}[2]{{\scriptsize[#1,\,#2]}}
\newcommand{\ooook}{OOOK-Eval}

\title{What Is in a WER? Reproducible Evaluation of Whisper for Sudanese Arabic}

\newcommand{\releaseurl}{\url{https://github.com/Saifuddin2Ahmed/sudanese-asr-eval}}

\author{Saifuddin Ahmed Abdelgadir Hamed \\
  Birtal Lab, Kampala, Uganda \\
  \texttt{saifuddin@birtallab.org} \\
  ORCID: \href{https://orcid.org/0009-0007-9719-7258}{0009-0007-9719-7258}}

\begin{document}
\maketitle

\begin{abstract}
Published word error rates (WER) for dialectal Arabic speech recognition are hard to compare, because text normalisation, decoding settings and data splits often go unreported. For practitioners who choose a model for real users, this matters. We re-evaluate Whisper on Sudanese Arabic with a fully specified protocol. (1)~An independent run of a public fine-tuned checkpoint reproduces its reported score (67.38 vs.\ 67.7 WER). (2)~Arabic orthographic normalisation (alef, ya, ta marbuta and the conjunction \emph{wa}) lowers cross-corpus WER by 10.8--12.7 points for models fine-tuned on the open Sudanese corpus, against 4.9--7.1 points for zero-shot models. Much of the apparent cross-corpus error is spelling convention, not recognition. (3)~Zero-shot Whisper large-v3 matches the fine-tuned small model on WER, though not on CER. (4)~On a new speaker-disjoint split, LoRA adaptation helps unseen speakers of the same corpus but brings no significant gain on a second Sudanese test set. (5)~A widely used mirror of the corpus pairs audio with the wrong transcripts: only 85 of 3,547 rows are aligned. We release the split, the normaliser, the scoring code and an alignment checker, and we turn these findings into an evaluation checklist for low-resource dialect ASR.
\end{abstract}

\section{Introduction}

Sudanese Arabic, an African variety of Arabic, is spoken by tens of millions of people in Sudan and in a large diaspora, but it has very few open speech resources. The main open corpus is about four hours of television speech \citep{mansour2022sdn}, released with a CNN-CTC baseline by \citet{mansour2022e2e}. \citet{mansour2026doing} recently fine-tuned Whisper \citep{radford2023whisper} on it, added pseudo-labels and synthetic speech, and introduced a new evaluation set, \ooook{}, drawn from the Sudanese portion of ADI17 \citep{shon2020adi17}.

Consider a small team building a voice feature for Sudanese users. It will pick a model, or decide whether to fine-tune one, from published WER. For a language variety this small, one number carries a lot of weight, and several hidden choices can move it by many points: how text is normalised before scoring, how the language token and decoding are set, and whether the test speakers were seen in training. Dialectal Arabic also has no standard orthography \citep{habash-etal-2012-conventional}, so two annotators can spell the same word differently, and WER counts every such difference as a full error \citep{ali2015mrwer}.

This paper does not propose a better model. It asks what the existing numbers measure and what a practitioner should take from them. Our contributions are:
\begin{enumerate}\itemsep0pt
\item an independent \textbf{reproduction} of the public 2022 checkpoint of \citet{mansour2022e2e}: 67.38 WER on \ooook{}, against 67.7 reported by \citet{mansour2026doing};
\item a \textbf{normalisation analysis} with three text views and a cumulative ablation. For models fine-tuned on the open corpus, normalisation explains 10.8--12.7 WER points on \ooook{}, mostly through alef and conjunction spelling;
\item a previously unreported \textbf{zero-shot baseline}: Whisper large-v3 matches the published fine-tuned small model on WER;
\item a \textbf{speaker-disjoint split} of the open corpus with dev-only model selection, and evidence that adaptation gains on unseen speakers do not transfer to a second Sudanese source;
\item an \textbf{error analysis}: rare repetition loops cause a large share of errors;
\item a \textbf{data-quality finding}: a popular mirror of the corpus has transcripts shifted against the audio;
\item \textbf{practical artefacts}: a documented normaliser and scorer, the split, a mirror-alignment checker and an evaluation checklist (\S\ref{sec:practice}).
\end{enumerate}

\section{Related Work}

\paragraph{Arabic dialect and African ASR.} The MGB-3 and MGB-5 challenges evaluated Egyptian and Moroccan ASR on broadcast and YouTube speech \citep{ali2017mgb3,ali2019mgb5}, and ADI17 collected dialectal broadcast audio from 17 countries \citep{shon2020adi17}. Recent resources include Casablanca, which covers eight dialects \citep{talafha-etal-2024-casablanca}, and SADA for Saudi Arabic \citep{alharbi2024sada}. \citet{talafha2023nshot} benchmarked Whisper on diverse Arabic varieties, and \citet{wang2025leaderboard} proposed a common leaderboard for Arabic ASR. Beyond Arabic, AfriSpeech-200 provides a pan-African benchmark of accented speech \citep{olatunji-etal-2023-afrispeech}, and participatory work in the Masakhane community stresses evaluation that is grounded in the communities served \citep{nekoto-etal-2020-participatory}. To our knowledge, the only Sudanese-specific ASR work is that of \citet{mansour2022e2e}, who built the corpus and a CNN-CTC model, and \citet{mansour2026doing}, who fine-tuned Whisper and introduced \ooook{}.

\paragraph{Normalisation and evaluation.} Dialectal Arabic lacks a standard spelling; CODA proposes conventions for computational use \citep{habash-etal-2012-conventional}. \citet{ali2015mrwer} showed that WER against a single reference greatly overstates errors for such languages and proposed multi-reference WER. Whisper ships a text normaliser because formatting differences inflate WER \citep{radford2023whisper}, but its language-agnostic ``basic'' variant does not unify Arabic letter variants. \citet{bisani2004bootstrap} proposed bootstrap confidence intervals for ASR. \citet{koenecke2024careless} documented Whisper's tendency to hallucinate, including repeated or invented text.

\section{Data and Protocol}
\label{sec:protocol}

\paragraph{Corpus and split.} We use the original Zenodo release (v1.1, CC BY 4.0) of the Sudanese dialect speech corpus \citep{mansour2022sdn}: 3,549 transcribed clips (4.03\,h) from two TV programmes, with per-clip speaker IDs. The corpus has no official split, and earlier work used a random one. We drop 182 clips without a speaker ID and one clip longer than 30\,s. We then split the remaining 3,366 clips by speaker with a fixed seed:
\begin{itemize}\itemsep0pt
\item \textbf{train}: 14 speakers, 2,695 clips, 3.01\,h;
\item \textbf{dev}: 15 speakers, 302 clips, 0.39\,h;
\item \textbf{test}: 5 speakers, 369 clips, 0.47\,h and 3,231 words, covering both programmes and both genders.
\end{itemize}
We call this test set the \emph{SDN test}.\footnote{Version~2 of this paper corrects all SDN-test numbers. In version~1, 58 of the 369 test clips, stored as anti-phase stereo, were reduced to near silence when the two channels were averaged. The main conclusions are unchanged; see Appendix~\ref{app:v2}.}

\paragraph{Second test set.} \ooook{} \citep{mansour2026doing} is a transcribed subset of the Sudanese ADI17 broadcast audio. We use its Hugging Face release (\href{https://huggingface.co/datasets/AymanMansour/OOOK-Eval-Final}{\nolinkurl{AymanMansour/OOOK-Eval-Final}}) at a pinned revision: 797 clips, 1.81\,h and 14,787 words. Its source and annotators differ from those of the training corpus. It has no stated licence, so we use it for evaluation only.

\paragraph{Systems.} We evaluate three groups of systems:
\begin{itemize}\itemsep0pt
\item zero-shot Whisper small, medium, large-v2, large-v3 and large-v3-turbo;
\item the 2022 public checkpoint of \citet{mansour2022e2e},\footnote{\url{https://huggingface.co/AymanMansour/Whisper-Sudanese-Dialect-small}} which its author trained on a random split of the corpus plus MSA. It may have seen our test speakers, so we score it on \ooook{} only;
\item two models we adapt on our train split: full fine-tuning of Whisper small, and LoRA \citep{hu2022lora} on large-v3-turbo.
\end{itemize}
Checkpoints are selected on dev WER only, and each test set is decoded once. Hyper-parameters are in Appendix~\ref{app:repro}.

\paragraph{Decoding and scoring.} All systems use greedy decoding, the Arabic language token, at most 225 new tokens and fp16. These settings follow the public fine-tuning template behind the 2022 checkpoint. We report corpus-level WER and CER in three text views:
\begin{itemize}\itemsep0pt
\item \emph{raw}: Unicode NFC and whitespace only;
\item \emph{basic}: Whisper's \texttt{BasicTextNormalizer}, as used by that template;
\item \emph{norm}: our Arabic normaliser.
\end{itemize}
Beyond \emph{basic}, the \emph{norm} view removes diacritics and tatweel, maps the alef variants \arimg{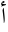}, \arimg{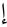} and \arimg{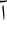} to \arimg{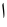}, unifies alef maqsura and hamza seats, maps ta marbuta \arimg{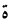} to \arimg{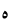}, and attaches a free-standing conjunction \arimg{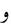} to the next word. It is applied identically to references and hypotheses. Confidence intervals are 95\% utterance-level bootstrap intervals with 1,000 resamples, and system comparisons use paired bootstrap on the same resamples.

\begin{table*}[t]
\centering
\small
\setlength{\tabcolsep}{2.6pt}
\begin{tabular}{l ccc c ccc c}
\toprule
 & \multicolumn{3}{c}{\ooook{} WER} & & \multicolumn{3}{c}{SDN speaker-disjoint test WER} & \\
\cmidrule(lr){2-4}\cmidrule(lr){6-8}
System & raw & basic & norm & CER norm & raw & basic & norm & CER norm \\
\midrule
small, zero-shot & 90.1\,\ci{83.8}{97.3} & 89.6 & 85.2\,\ci{79.0}{92.5} & 51.2 & 92.0\,\ci{81.1}{104.2} & 91.3 & 87.7\,\ci{75.9}{100.5} & 47.3 \\
medium, zero-shot & 72.4\,\ci{67.6}{77.7} & 71.6 & 66.3\,\ci{61.2}{71.7} & 37.0 & 79.4\,\ci{68.8}{95.6} & 78.9 & 72.1\,\ci{60.6}{89.9} & 44.0 \\
large-v2, zero-shot & 69.4\,\ci{65.4}{74.2} & 68.5 & 62.5\,\ci{58.2}{67.5} & 36.3 & 68.1\,\ci{63.4}{75.2} & 67.4 & 58.4\,\ci{52.9}{66.4} & 31.1 \\
\quad auto language ID & 70.3\,\ci{66.0}{75.4} & 69.6 & 63.4\,\ci{59.1}{68.7} & 36.9 & -- & -- & -- & -- \\
large-v3, zero-shot & \textbf{66.4}\,\ci{60.8}{73.0} & 65.5 & 59.3\,\ci{53.6}{66.0} & 31.5 & 76.0\,\ci{64.7}{94.3} & 78.2 & 67.9\,\ci{55.4}{87.7} & 34.7 \\
turbo, zero-shot & 70.1\,\ci{64.2}{77.3} & 69.7 & 63.2\,\ci{57.0}{70.5} & 34.4 & 83.2\,\ci{68.6}{100.3} & 82.7 & 75.5\,\ci{59.8}{94.0} & 38.1 \\
\midrule
2022 public small ckpt & 67.4\,\ci{65.9}{69.0} & 67.4 & \textbf{56.5}\,\ci{54.8}{58.3} & \textbf{24.6} & -- & -- & -- & -- \\
small, full FT (ours) & 75.8\,\ci{72.0}{80.0} & 75.5 & 65.1\,\ci{61.4}{69.2} & 33.1 & 60.2\,\ci{52.9}{69.2} & 58.7 & 58.2\,\ci{52.1}{65.6} & 27.0 \\
turbo + LoRA (ours) & 71.5\,\ci{67.0}{76.7} & 71.5 & 58.8\,\ci{54.4}{64.1} & 27.0 & \textbf{43.0}\,\ci{40.3}{45.6} & 42.9 & \textbf{43.5}\,\ci{40.8}{46.1} & \textbf{15.9} \\
\bottomrule
\end{tabular}
\caption{WER and CER (\%) with 95\% bootstrap intervals. \emph{raw}: whitespace only; \emph{basic}: Whisper \texttt{BasicTextNormalizer}; \emph{norm}: our Arabic normaliser; turbo: large-v3-turbo. Greedy decoding, with Arabic forced unless noted. The 2022 checkpoint was trained on a random split that may include our test speakers, so we do not score it on the SDN test. On \ooook{}, \citet{mansour2026doing} reports 109, 84.3 and 78.8 WER for zero-shot small, medium and large-v2, and 67.7 for the 2022 small model. The normalisation and decoding settings behind those numbers are not stated.}
\label{tab:main}
\end{table*}

\section{Results}

Table~\ref{tab:main} gives all scores, and Figure~\ref{fig:wer} in the appendix plots them.

\paragraph{Reproduction.} On \ooook{}, the public 2022 small checkpoint scores 67.38 raw WER and 27.28 raw CER, against 67.7 and 27.7 reported by \citet{mansour2026doing}. Raw and basic WER differ by only 0.01, so the published number is most plausibly a raw or basic score. Our pipeline matches it within 0.5 points, and this match anchors the other comparisons.

Our zero-shot scores are lower than the published zero-shot rows: large-v2 scores 69.4 here, against 78.8 reported. Language identification does not explain the gap. Letting Whisper detect the language costs only 0.9 points (paired $\Delta$ $-0.9$ \ci{$-2.7$}{$+0.5$}). Since the decoding and segmentation settings for those rows are not published, we cannot locate the difference. We report it only so that later work knows which protocol produced which number.

\begin{table}[t]
\centering
\small
\setlength{\tabcolsep}{3pt}
\begin{tabular}{lcccccc}
\toprule
System & raw & +p/d & +alef & +ya & +ta & +wa \\
\midrule
\multicolumn{7}{l}{\emph{\ooook{}}} \\
large-v3 ZS & 66.4 & 65.3 & 61.9 & 61.6 & 60.0 & 59.3 \\
turbo ZS & 70.1 & 69.3 & 66.0 & 65.6 & 64.1 & 63.2 \\
2022 ckpt & 67.4 & 67.4 & 64.1 & 62.6 & 61.4 & 56.5 \\
small FT & 75.8 & 75.5 & 72.7 & 71.9 & 70.5 & 65.1 \\
turbo LoRA & 71.5 & 71.5 & 68.2 & 66.2 & 64.8 & 58.8 \\
\midrule
\multicolumn{7}{l}{\emph{SDN test}} \\
large-v3 ZS & 76.0 & 74.8 & 73.6 & 72.4 & 72.3 & 67.9 \\
turbo ZS & 83.2 & 82.6 & 81.0 & 79.9 & 79.8 & 75.5 \\
small FT & 60.2 & 58.7 & 58.5 & 57.4 & 57.3 & 58.2 \\
turbo LoRA & 43.0 & 42.9 & 42.7 & 42.0 & 41.9 & 43.5 \\
\bottomrule
\end{tabular}
\caption{Cumulative WER (\%) as normalisation steps are added. The steps are punctuation and diacritics (p/d), alef variants, ya and hamza seats, ta marbuta, and attaching the conjunction \arimg{ar07} (wa); the last column equals \emph{norm}. ZS: zero-shot. On the SDN test, attaching \arimg{ar07} slightly raises WER for the fine-tuned models: these models already segment words as the references do, and the merge shortens the reference.}
\label{tab:ablation}
\end{table}

\paragraph{Normalisation.} On \ooook{}, moving from raw to norm lowers WER by 4.9--7.1 points for zero-shot models. The drop is 10.9 for the 2022 checkpoint, 10.8 for our full fine-tune and 12.7 for our LoRA model. Whisper's basic normaliser recovers at most one point of this.

The cumulative ablation in Table~\ref{tab:ablation} shows where the gain comes from. On \ooook{}, alef unification gives about 3 points for every system. Attaching the conjunction \arimg{ar07} gives 4.9--5.9 points for the three models fine-tuned on the corpus, but under one point for zero-shot models. The reason is a difference in spelling conventions. The training corpus writes \arimg{ar01}/\arimg{ar02}, a detached \arimg{ar07} and \arimg{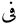}, while \ooook{} annotators mostly write bare \arimg{ar04} and an attached \arimg{ar07}. Fine-tuned models learn the first convention, and raw WER then penalises them on the second corpus for spelling, not recognition. On the SDN test, where training and test data share a convention, normalisation changes fine-tuned WER by about two points or less (raw minus norm: 2.0 for the full fine-tune, $-0.5$ for LoRA).

Single-view WER can therefore change conclusions. Under raw WER, our LoRA model is 5.2 points behind zero-shot large-v3 on \ooook{}; under norm it is 0.4 points ahead. Neither difference is significant. Its gap between the two test sets also narrows from 28.6 points under raw WER (71.5 vs.\ 43.0) to 15.3 under norm (58.8 vs.\ 43.5).

\paragraph{A strong zero-shot baseline.} Zero-shot large-v3 reaches 66.39 raw WER on \ooook{}. This is statistically indistinguishable from the fine-tuned 2022 checkpoint (paired $\Delta$ $-1.0$ \ci{$-6.6$}{$+5.3$}). The fine-tuned model remains clearly better on CER (27.3 vs.\ 33.6 raw) and produces no runaway outputs, so fine-tuning does help. Large-v3 is nevertheless the natural zero-shot reference for future comparisons.

\paragraph{Speaker-disjoint adaptation.} On unseen speakers of the same corpus, LoRA reduces large-v3-turbo WER from 83.2 to 43.0 (paired $\Delta$ $-40.2$ \ci{$-56.5$}{$-26.8$}). Under norm the reduction is $-32.0$ \ci{$-49.7$}{$-17.1$}, also significant. LoRA is also better than our full fine-tune of small on this test set ($-17.2$ \ci{$-26.4$}{$-10.2$} raw). On \ooook{}, the same model shows no significant change: $+1.4$ \ci{$-6.7$}{$+8.6$} raw and $-4.4$ \ci{$-12.2$}{$+2.9$} norm. Three hours of TV speech thus help on new speakers of the same programmes, but we find no evidence that they help on another Sudanese source.

Our plain full fine-tune of Whisper small is 8.5 points \ci{4.9}{12.9} worse than the 2022 checkpoint on \ooook{}. It differs in several ways: less data (3.01\,h, no MSA), a smaller batch and a noisy 0.39\,h dev set. We cannot separate these causes.

\paragraph{Error analysis.} Clips whose hypothesis is more than three times the reference length are rare, at most 2\% of clips for large models, but costly. On \ooook{} they hold 14\% of all word errors for zero-shot large-v3 and 16\% for turbo. On the SDN test, 23\% of zero-shot turbo's word errors come from 6 of 369 clips, while the LoRA model produces no such clip. Typical failures are loops (\arimg{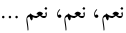}) and stock phrases such as \arimg{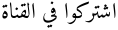} (``subscribe to the channel''). The 2022 checkpoint produces none. Zero-shot models also replace Sudanese forms with MSA ones: for example, the object suffix \arimg{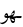} becomes \arimg{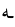}. Fine-tuning corrects this (Appendix~\ref{app:examples}).

As an \emph{exploratory} check, we applied a reference-free rule that collapses three or more repeats of an n-gram. It lowers LoRA's \ooook{} norm WER to 53.1 and zero-shot large-v3 raw WER to 58.0. We chose this rule after seeing the outputs, so these figures are not results. They only show that repetition control at decoding time deserves a pre-registered test.

\paragraph{A misaligned mirror.} The Hugging Face dataset \href{https://huggingface.co/datasets/arbml/sudanese_dialect_speech}{\nolinkurl{arbml/sudanese_dialect_speech}} is a convenient mirror of the corpus, but most of its rows pair an audio file with the transcript of a neighbouring clip. Only 85 of its 3,547 rows match the Zenodo transcript of the same file; 1,554 match the next clip and 1,907 the clip after it. This pattern is consistent with an ordering offset introduced during conversion. Models trained or scored on this mirror would learn, or be judged on, the wrong text. All our experiments load the original Zenodo archive, verified by MD5.

\section{From Research to Practice}
\label{sec:practice}

The workshop theme asks how research reaches real users. For low-resource dialect ASR, our results suggest that the weakest link is often evaluation, not modelling. This section turns them into concrete guidance and tools.

\paragraph{An evaluation checklist.} For dialectal Arabic ASR, and plausibly for other African language varieties without a settled orthography, we suggest the following:
\begin{enumerate}\itemsep0pt
\item Report WER in at least two text views, raw and a documented normalisation, and release the normaliser.
\item State decoding settings: search, language token, maximum length and repetition control.
\item Split by speaker, and select checkpoints on a dev set, never on test.
\item Add a second test source to measure transfer beyond one corpus.
\item Give confidence intervals, paired tests and test-set speaker counts.
\item Include current zero-shot models, such as large-v3, as baselines.
\item Report CER and hallucination counts next to WER.
\item Load data from the original release, and check audio--text alignment before training.
\end{enumerate}
Each item is cheap. Items 1, 2 and 8 address the three largest sources of discrepancy we found: spelling conventions, unstated decoding settings and misaligned data.

\paragraph{Released tools.} We release a small, dependency-light Python package and command-line tool. It implements the three text views, with unit tests and a parity check against the \texttt{BasicTextNormalizer} of Whisper. It also computes corpus-level WER and CER with bootstrap intervals and paired comparisons between two systems. Alongside it we release:
\begin{itemize}\itemsep0pt
\item the speaker-disjoint split, as clip and speaker IDs only;
\item text-free per-clip error counts for both test sets, and a script that recomputes every number in this paper from them on a CPU;
\item the mirror-alignment checker.
\end{itemize}
The checker uses only the Python standard library. It reads the mirror's rows through the Hugging Face dataset viewer API, without downloading audio. It reads the original transcripts by HTTP range requests into the Zenodo archive, so the 1.45\,GB file is not downloaded either. It then tests each row against the transcript of the same file and of nearby files. The same check can be adapted to any mirror of a speech corpus that keeps original file names.

\paragraph{What this means for model choice.} Three practical points follow for a team deploying Sudanese ASR today.
\begin{itemize}\itemsep0pt
\item \emph{Zero-shot large-v3 is competitive on WER but needs guardrails.} About 1.4\% of its \ooook{} outputs are runaway loops or stock phrases, and they cause 14\% of its word errors. A length or repetition check is a simple safeguard before showing output to users.
\item \emph{The fine-tuned small checkpoint is a strong low-compute option.} It has about 0.24B parameters, against 1.55B for large-v3 \citep{radford2023whisper}, the best CER on \ooook{} and no runaway outputs. This makes it the more practical candidate for on-device or low-bandwidth use, although we did not measure latency or memory.
\item \emph{Adaptation is affordable, but its gains may not travel.} LoRA training took under 7\,h on a single free-tier 16\,GB GPU, within reach of small labs. Its gain did not transfer to a second source, so a deployment claim needs test data from the target domain and speakers. A larger, multi-source and speaker-diverse Sudanese test set is the most useful next resource.
\end{itemize}

\section{Conclusion}

A careful re-evaluation of Whisper on Sudanese Arabic shows four things. The published fine-tuned result is reproducible. Much of the apparent gap between corpora is spelling convention rather than recognition error. A modern zero-shot model is a strong baseline. Adaptation on three hours of speech helps unseen speakers of the same corpus, but not a second source. None of these findings needed new data or large compute; they needed a fully specified protocol. We hope the released normaliser, split and alignment checker make such protocols the default for Sudanese and other low-resource dialect ASR.

\section*{Limitations}

\begin{itemize}\itemsep0pt
\item \textbf{Small, unbalanced test set.} The speaker-disjoint test set has only 5 speakers, and 3 of them contribute 87\% of its words. Per-speaker LoRA WER ranges from 35\% to 60\%. Utterance-level bootstrap intervals therefore understate speaker-level uncertainty, and SDN-test differences under about 10 points should be treated as unresolved.
\item \textbf{Noisy model selection.} The dev set (0.39\,h) is small, which made checkpoint selection noisy.
\item \textbf{Narrow coverage.} All training data come from two TV programmes in the Khartoum dialect. Our results say little about other Sudanese varieties, spontaneous conversation or telephone speech.
\item \textbf{One normaliser.} Our normaliser is one set of choices. Attaching \arimg{ar07} can merge words that should stay separate, as the SDN-test ablation shows, and we did not compute multi-reference WER.
\item \textbf{No user evaluation.} We did not evaluate with real users or on a downstream task. Our practical recommendations rest on offline metrics.
\item \textbf{Limited runs.} Each fine-tuning run used one seed and one hyper-parameter setting, with greedy decoding only.
\item \textbf{Indicative comparisons.} The normalisation, decoding and checkpoint-selection settings of \citet{mansour2026doing} are not published. Apart from the reproduced checkpoint, comparisons with the reported numbers are indicative only.
\item \textbf{Post-hoc figures.} The loop-collapse figures are post hoc.
\item \textbf{No retraining after the v2 correction.} Our two adapted models were trained before the stereo fix (Appendix~\ref{app:v2}): 57 of their 2,695 training clips were near-silent but kept their full transcripts, and 115 of the 302 dev clips used for checkpoint selection were affected. Retrained models may score differently on both test sets. The v2 hypotheses for the 58 affected test clips were decoded on a CPU in fp32, not on the original GPU.
\end{itemize}

\section*{Ethical Considerations}

We use only previously published data:
\begin{itemize}\itemsep0pt
\item the Sudanese dialect speech corpus, released under CC BY 4.0 and derived from publicly broadcast television programmes;
\item \ooook{}, which has no stated licence, so we use it for evaluation only and do not redistribute its audio or transcripts.
\end{itemize}
We collected no new recordings and no new personal data. Speaker IDs are the corpus's anonymous numbers. We release only derived artefacts: split lists, code, SDN-test hypotheses and text-free per-clip counts.

Before publication, we reported the mirror misalignment privately to the corpus author and to the mirror's maintainers, so that it can be fixed at the source.

Speech recognition for under-served varieties can widen access, but it can also enable surveillance. The models here are research baselines, with error rates far too high for unsupervised use.

\ifcamera
\section*{Acknowledgements}

We thank Ayman Mansour for releasing the Sudanese dialect speech corpus, the \ooook{} evaluation set and his fine-tuned checkpoints; this work would not exist without them.
\fi

\bibliography{refs}

\appendix

\section{Reproducibility}
\label{app:repro}

\paragraph{Release.} The release contains:
\begin{itemize}\itemsep0pt
\item the split file (clip IDs and speaker IDs only) and the normaliser with its unit tests;
\item the scoring and bootstrap code, the training configurations and the notebooks;
\item all SDN-test hypotheses, and the per-clip error counts for \ooook{} without its text.
\end{itemize}
\ifcamera
The code is under Apache-2.0, and the split and scores are under CC BY 4.0. The release is available at \releaseurl{} and archived on Zenodo (\href{https://doi.org/10.5281/zenodo.24354751}{doi:10.5281/zenodo.24354751}, all versions).
\else
For review, it is provided as anonymised supplementary material: one software archive and one data archive. Code is under Apache-2.0, and the split and scores under CC BY 4.0. It will be released publicly on acceptance.
\fi
A single script recomputes every number in this paper from the released counts.

\paragraph{Compute.} All training and original decoding runs used one NVIDIA T4 (16\,GB) on Kaggle, in fp16, for about 10 GPU-hours in total: zero-shot decoding about 1\,h, full fine-tuning 1.4\,h and LoRA 6.8\,h. Software: Transformers 4.56.2 \citep{wolf-etal-2020-transformers}, PEFT 0.17.1 and jiwer 3.1.0. A full \texttt{pip freeze} is saved with every run. On a CPU in fp32, our scores differ by at most 0.5 WER. The version~2 re-decoding of the 58 affected SDN-test clips ran on a CPU (Appendix~\ref{app:v2}).

\paragraph{Seeds and data.} The split seed is 354; the training seed and the bootstrap seed are both 42, with 1,000 resamples. The Zenodo archive is checked by MD5, and \ooook{} is loaded at revision \texttt{e256ab5}. Audio is down-mixed to mono (750 of the 3,550 Zenodo WAV files are stereo) and resampled to 16\,kHz. Since version~2, stereo clips whose channels are in anti-phase are down-mixed as $(L-R)/2$ and all others as $(L+R)/2$ (Appendix~\ref{app:v2}).

\paragraph{Hyper-parameters.} \emph{Full fine-tune (small):}
\begin{itemize}\itemsep0pt
\item learning rate $10^{-5}$, 300 warm-up steps and effective batch 16;
\item up to 4,000 steps, with dev evaluation every 250 steps and early stopping after 4 evaluations without gain;
\item best dev step 1,500.
\end{itemize}
\emph{LoRA (large-v3-turbo):}
\begin{itemize}\itemsep0pt
\item rank 32, $\alpha=64$ and dropout 0.05, applied to the attention projections and feed-forward layers;
\item learning rate $5\cdot10^{-4}$, 100 warm-up steps and effective batch 16;
\item up to 2,000 steps, with evaluation every 200 steps and gradient checkpointing;
\item best dev step 1,000.
\end{itemize}
The selection metric for both is dev WER in the norm view.

\section{Changes in Version 2}
\label{app:v2}

\paragraph{The issue.} 750 of the WAV files in the corpus are stereo. In 230 clips of our split (58 test, 115 dev and 57 train clips), one channel is a sign-inverted copy of the other: the correlation between the two channels lies between $-0.81$ and $-0.996$. Version~1 down-mixed every stereo clip by averaging its channels, which cancels such a clip to near silence. We found the issue when a commercial recogniser returned empty transcripts for exactly these test clips on our pre-processed audio but not on the original files. The channel correlation then identified the 230 clips; every other stereo clip in the split has a correlation of at least 0.5. All 58 affected test clips come from one episode of one programme (speakers 17, 20 and 26).

\paragraph{What changed.} The loader now down-mixes anti-phase clips as $(L-R)/2$. For every system scored on the SDN test, we re-decoded only the 58 affected clips, with the same checkpoints and decoding settings, and kept the other 311 hypotheses. This re-decoding ran on a CPU in fp32. As a check, we re-decoded 96 unaffected control clips in the same way (16 per system; 8 each for large-v2 and large-v3): 93 hypotheses were identical to the original GPU ones, and the control WER of each system moved by at most 0.8 points. All SDN-test numbers in Tables~\ref{tab:main} and~\ref{tab:ablation}, Figure~\ref{fig:wer}, the paired tests and the error analysis were recomputed with the same code. \ooook{} scores are unchanged; one \ooook{} entry that version~1 had rounded up from 69.45 (zero-shot large-v2, raw WER) now reads 69.4. Table~\ref{tab:v2} compares the SDN-test WER of both versions. The release documents the fix and keeps the version~1 behaviour available for reproduction.

\paragraph{Effect on the findings.} The correction lowers raw SDN-test WER for six of the seven systems, most for systems that had looped on the near-silent clips (zero-shot small, 136.4 to 92.0; zero-shot medium, 98.3 to 79.4; LoRA, 59.9 to 43.0). Zero-shot large-v2 and large-v3 improve by 7--8 points (76.0 to 68.1 and 83.0 to 76.0). Only zero-shot turbo gets worse (80.2 to 83.2). On the near-silent clips this model had produced short outputs (388 words against 555 reference words), and on the restored audio it produces longer ones (838 words), including some loops. The LoRA model's gains over zero-shot turbo and over our full fine-tune of small on unseen speakers of the corpus are now significant under both text views. Under norm, zero-shot large-v2 (58.4) is now close to our full fine-tune of small (58.2) on the SDN test. The findings on \ooook{}, including the lack of a significant LoRA gain there, are unchanged, so the contrast between the two test sets is stronger than in version~1. The SDN-test hallucination example of version~1 was an affected clip and has been replaced. The adapted models were not retrained (see Limitations).

\begin{table}[h]
\centering
\small
\begin{tabular}{lcccc}
\toprule
 & \multicolumn{2}{c}{WER raw} & \multicolumn{2}{c}{WER norm} \\
\cmidrule(lr){2-3}\cmidrule(lr){4-5}
System & v1 & v2 & v1 & v2 \\
\midrule
small ZS & 136.4 & 92.0 & 136.6 & 87.7 \\
medium ZS & 98.3 & 79.4 & 93.2 & 72.1 \\
large-v2 ZS & 76.0 & 68.1 & 67.7 & 58.4 \\
large-v3 ZS & 83.0 & 76.0 & 76.5 & 67.9 \\
large-v3-turbo ZS & 80.2 & 83.2 & 72.9 & 75.5 \\
small, full FT & 65.3 & 60.2 & 64.8 & 58.2 \\
large-v3-turbo + LoRA & 59.9 & 43.0 & 61.7 & 43.5 \\
\bottomrule
\end{tabular}
\caption{SDN-test WER (\%) in version~1 (58 clips near-silent) and version~2 (corrected).}
\label{tab:v2}
\end{table}

\section{Examples}
\label{app:examples}

All examples in Table~\ref{tab:examples} are from the SDN test set (CC BY 4.0) and are given with their clip IDs.

\begin{table}[h]
\centering
\small
\renewcommand{\arraystretch}{1.25}
\begin{tabular}{@{}l>{\raggedleft\arraybackslash}p{0.72\columnwidth}@{}}
\toprule
\multicolumn{2}{@{}l}{\emph{Dialect morphology} (\texttt{hm\_25\_0049})} \\
Ref. & \arimg{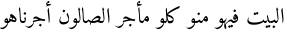} \\
LoRA & \arimg{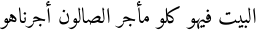} \\
v3 ZS & \arimg{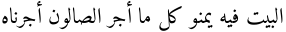} \\
\midrule
\multicolumn{2}{@{}l}{\emph{Hallucination} (\texttt{wb\_01\_0187})} \\
Ref. & \arimg{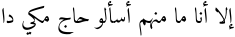} \\
v3 ZS & \arimg{ar10} \\
turbo ZS & \arimg{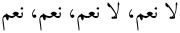} \ldots{} (77 words) \\
\midrule
\multicolumn{2}{@{}l}{\emph{Genuine recognition error} (\texttt{hm\_02\_0074})} \\
Ref. & \arimg{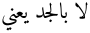} \\
LoRA & \arimg{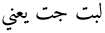} \\
\bottomrule
\end{tabular}
\caption{Qualitative examples. First: the zero-shot model turns the Sudanese object suffix \arimg{ar11} into MSA \arimg{ar12}, and \arimg{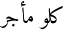} into \arimg{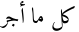}; each change counts as a full word error. Second: zero-shot large-v3 outputs a stock phrase (``subscribe to the channel''), and zero-shot turbo loops. Third: a fast, reduced pronunciation that no normalisation can repair.}
\label{tab:examples}
\end{table}

\begin{figure*}[t]
\centering
\includegraphics[width=\textwidth]{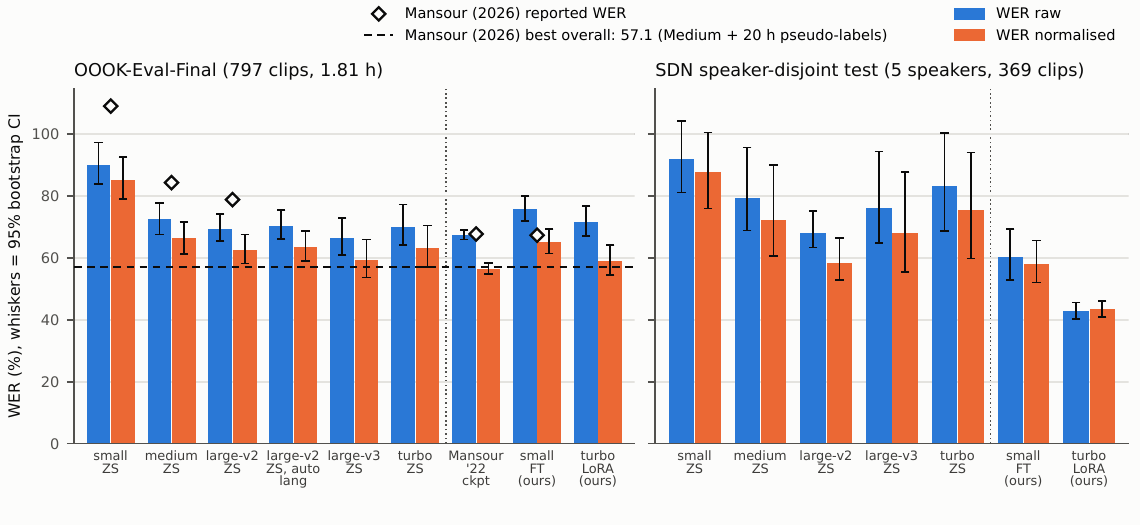}
\caption{WER per system, raw (blue) and normalised (orange), with 95\% bootstrap intervals. Diamonds show the WER reported by \citet{mansour2026doing}. The dashed line shows his best overall result (57.1, Whisper medium with about 20\,h of pseudo-labels), which is not a like-for-like comparison.}
\label{fig:wer}
\end{figure*}

\end{document}